# Enhanced Photovoltaic Power Forecasting: An iTransformer and LSTM-Based Model Integrating Temporal and Covariate Interactions

Guang Wu
*School of Automation*
*Central South University*
Changsha, China
wuguangcsuer@csu.edu.cn

Yun Wang*
*School of Automation*
*Central South University*
Changsha, China
wangyun19@csu.edu.cn

Qian Zhou
*School of Automation*
*Central South University*
Changsha, China
224611074@csu.edu.cn

Ziyang Zhang
*School of Automation*
*Central South University*
Changsha, China
234612221@csu.edu.cn

***Abstract*—Accurate photovoltaic (PV) power forecasting is critical for integrating renewable energy sources into the grid, optimizing real-time energy management, and ensuring energy reliability amidst increasing demand. However, existing models often struggle with effectively capturing the complex relationships between target variables and covariates, as well as the interactions between temporal dynamics and multivariate data, leading to suboptimal forecasting accuracy. To address these challenges, we propose a novel model architecture that leverages the iTransformer for feature extraction from target variables and employs long short-term memory (LSTM) to extract features from covariates. A cross-attention mechanism is integrated to fuse the outputs of both models, followed by a Kolmogorov–Arnold network (KAN) mapping for enhanced representation. The effectiveness of the proposed model is validated using publicly available datasets from Australia, with experiments conducted across four seasons. Results demonstrate that the proposed model effectively capture seasonal variations in PV power generation and improve forecasting accuracy. Code is available at this repository: https://github.com/laowu-code/iTansformer_LSTM_CA_KAN.**

## I. INTRODUCTION

Against the backdrop of a global energy crisis, there is an increasing focus worldwide on energy conservation, emissions reduction, and achieving carbon neutrality [1]. Photovoltaic (PV) power generation, as a clean and renewable energy source, is instrumental in advancing the energy transition. However, PV generation is significantly impacted by seasonal variations, making it challenging to achieve high forecasting accuracy. Moreover, the intermittent and fluctuating nature of PV generation presents substantial challenges for grid integration [2]. Therefore, accurate PV power forecasting is crucial for the successful integration of solar energy into the grid and for optimizing real-time grid scheduling.

### *A. Literature review*

Many models have been applied to PV power forecasting, which can be categorized into three main types based on their structure: physical models, statistical models, and machine learning models.

Physical models rely solely on primary design parameters of PV systems and numerical weather prediction (NWP), requiring no historical data. This makes them widely applicable in PV power stations where data availability is limited. Reference [3] conducted an in-depth study on physical models for PV power forecasting using NWP data, providing a comprehensive analysis. Additionally, reference [4] developed an ensemble model chain driven by ensemble NWP data to generate probabilistic forecasts, which they validated on 14 utility-scale PV plants across Hungary. Similarly, reference [5] introduced a physical model that combines the persistence method, cloud motion vectors, and irradiance predictions from NWP. In a related approach, reference [6] applied a physical model incorporating DIRINT decomposition, PEREZ transposition, and a proprietary PV performance simulation, utilizing both NWP and on-site irradiance data for PV power prediction. However, the heavy reliance on NWP accuracy in physical models, along with error amplification throughout the physical modeling chain, restricts their applicability, especially in contexts requiring precise real-time forecasting.

Statistical models are widely used in PV power forecasting, especially when a considerable amount of historical data is available and the PV system operates under relatively stable conditions. These models, like autoregressive [7], autoregressive moving average [8], autoregressive integrated moving average (ARIMA) [9] and seasonal ARIMA [10], rely on historical data patterns and weather variables to capture trends and seasonality without requiring detailed physical system information. They are particularly effective in scenarios with long-term, continuous PV generation data, enabling them to detect relationships and trends that aid short-term and medium-term forecasting. However, statistical models face limitations in handling nonlinear relationships and adapting to rapidly changing weather conditions.

Machine learning models in PV power forecasting can be categorized into traditional machine learning and deep learning approaches, each with distinct strengths. Traditional machine learning models, such as linear regression [11], random forest [12] and support vector machine [13], are advantageous in scenarios where interpretability and quick training are essential, often applied to day-ahead forecasts with moderate data requirements. These models handle nonlinear relationships and moderate data sizes effectively, making them suitable for simpler forecasting applications. In contrast, deep learning models, including convolutional neural networks [14], temporal convolutional networks [15], long short-term memory (LSTM)

This work was supported in part by the National Natural Science Foundation of China under Grant 62376289, and in part by the Natural Science Foundation of Hunan Province, China under Grant 2024JJ4069. Corresponding author: Yun Wang.

[16], gated recurrent unit [17], Transformers [18] and Kolmogorov–Arnold networks (KANs) [19], excel in capturing complex patterns and dependencies within large, high-dimensional datasets, particularly for very-short-term and intra-day forecasts. The iTransformer model proposed by [20] restructures the Transformer architecture by embedding time points as individual variate tokens, enabling enhanced multivariate correlation capture and nonlinear representation, thereby improving time series forecasting performance.

### *B. Motivation and contributions*

Despite structural adjustments for long-term or short-term forecasts, the above models still faces limitations in multivariate-to-univariate forecasting tasks:

- They lack effective modeling and feature extraction between the target variable and covariates.
- They fail to adequately capture the interactions between temporal dynamics and multivariate relationships, which can hinder the accuracy of the forecasting outcomes.

Therefore, this study introduces a novel forecasting framework that synergistically integrates the iTransformer and LSTM architectures to enhance the accuracy of PV power forecasts. The model's architecture is designed to optimize feature extraction and interaction modeling across multiple dimensions. The iTransformer is utilized to extract features from the target variable, while the LSTM effectively processes covariates, allowing for a comprehensive analysis of the underlying data patterns. The integration is further refined through a cross-attention mechanism, followed by a KAN layer that maps the fused features to the desired output dimension. The main contributions of this study are summarized as follows:

- **Effective Multivariate Feature Extraction**: By leveraging both iTransformer and LSTM, the model improves feature extraction from the target variable and covariates, allowing for a more comprehensive understanding of the factors influencing PV power.
- **Robust Interaction Modeling**: The use of cross-attention enables the model to effectively capture interactions between temporal dynamics and multivariate relationships, enhancing the representation of how covariate changes affect the target variable over time.
- **Effectiveness Validation and Seasonal Analysis**: A case study utilizing publicly available datasets from Australia is conducted to assess the effectiveness of the proposed model. Experiments are performed across four seasons, demonstrating the model's robust performance and accuracy in capturing seasonal variations in PV power forecasting.

## II. Methodology and Model

This section introduces the proposed photovoltaic power forecasting model, which incorporates Cross-Attention, iTransformer, LSTM, and KAN.

### *A. Cross-Attention*

Unlike self-attention, in cross-attention, the queries $Q$ and keys/values $(K, V)$ derived from different modules. Given $\boldsymbol{V}_1 \in \mathbb{R}^{s_1 \times d_1}, \boldsymbol{V}_2 \in \mathbb{R}^{s_2 \times d_2}$ from these distinct modules,

$$\mathrm{CA}(\boldsymbol{V}_1, \boldsymbol{V}_2) = \mathrm{softmax}\left(\frac{(\boldsymbol{V}_1\boldsymbol{W}^Q)(\boldsymbol{V}_2\boldsymbol{W}^K)^\top}{\sqrt{d_k}}\right)(\boldsymbol{V}_2\boldsymbol{W}^V), \quad (1)$$

where $CA(\cdot)$ is the cross-attention operation, $softmax(\cdot)$ is the softmax function, $\boldsymbol{W}^Q \in \mathbb{R}^{d_1 \times d_k}$, $\boldsymbol{W}^K \in \mathbb{R}^{d_2 \times d_k}$ and $\boldsymbol{W}^V \in \mathbb{R}^{d_2 \times d_k}$ are the query, key and value projection matrices, respectively. When $\boldsymbol{V}_1$ and $\boldsymbol{V}_2$ are identical, cross-attention transforms into self-attention

$$\mathrm{SA}(\boldsymbol{V}_1) = \mathrm{CA}(\boldsymbol{V}_1, \boldsymbol{V}_1), \quad (2)$$

where $\mathrm{SA}(\cdot)$ is the self-attention operation.

### *B. iTransformer*

The Transformer architecture has proven to be highly effective in capturing long-term dependencies within time series data. The iTransformer, proposed in [20], embeds each time series as an independent variable token, rather than embedding different variables along the same time dimension. This method emphasizes capturing multivariate correlations and employs a feed-forward network for enhanced sequence representation learning. Given a sequence $\{\boldsymbol{X}_1, \boldsymbol{X}_2, \cdots, \boldsymbol{X}_m\} \in \mathbb{R}^{m \times l}$ with $m$ time variables and a lag length of $l$ to get the results $\{\widehat{\boldsymbol{Y}}_1, \widehat{\boldsymbol{Y}}_2, \cdots, \widehat{\boldsymbol{Y}}_m\} \in \mathbb{R}^{m \times p}$ with desired forecast length $p$, the iTransformer embeds each individual variable separately

$$\begin{aligned}
\boldsymbol{h}_i^0 &= \mathrm{Embedding}(\boldsymbol{X}_i), i = 1,2,\cdots,m, \\
\widehat{\boldsymbol{h}}_i^l &= \mathrm{LN}\left(\boldsymbol{h}_i^l + \mathrm{SA}\left(\boldsymbol{h}_i^l\right)\right), l = 0,1,\cdots,L-1, \\
\boldsymbol{h}_i^{l+1} &= \mathrm{LN}(\mathrm{FFN}(\widehat{\boldsymbol{h}}_i^l) + \widehat{\boldsymbol{h}}_i^l), l = 0,1,\cdots,L-1, \\
\widehat{\boldsymbol{Y}}_i &= \mathrm{Projection}(\boldsymbol{h}_i^L),
\end{aligned} \quad (3)$$

where $\mathrm{Embedding}(\cdot)$ embeds the sequence length $l$ into $d_h$, while the $\mathrm{Projection}(\cdot)$ projects $\boldsymbol{h}_i^L \in \mathbb{R}^{1 \times d_h}$ to the desired forecast $\widehat{Y}_i \in \mathbb{R}^{1 \times p}$ and $L$ is the layer of the iTransformer. Both operations are implemented through multi-layer perceptron (MLP). $\mathrm{LN}(\cdot)$ is the Layer Normalization operation and $\mathrm{FFN}(\cdot)$ denotes the feed forward network.

### *C. Long short-term memory*

LSTM effectively manages long sequential data by modeling long-term dependencies through its gating mechanism. The input gate, forget gate, and output gate control the flow of information, determining what to store, discard, or retrieve at each time step. The cell state $\boldsymbol{c}_t$ enables smooth propagation of information across time steps, reducing issues like gradient vanishing and explosion that are common in traditional RNNs. Its mathematical expressions are as follows:

$$\begin{aligned}
\boldsymbol{i}_t &= \sigma\left(\boldsymbol{W}_i[\boldsymbol{h}_{t-1}, \boldsymbol{x}_t] + b_i\right), \\
\boldsymbol{f}_t &= \sigma\left(\boldsymbol{W}_f[\boldsymbol{h}_{t-1}, \boldsymbol{x}_t] + b_f\right), \\
\boldsymbol{o}_t &= \sigma\left(\boldsymbol{W}_o[\boldsymbol{h}_{t-1}, \boldsymbol{x}_t] + b_o\right),
\end{aligned}$$

$$\boldsymbol{c}_t = \boldsymbol{f}_t \odot \boldsymbol{c}_{t-1} + \boldsymbol{i}_t \odot \tanh\left(\boldsymbol{W}_c\left[\boldsymbol{h}_{t-1}, \boldsymbol{x}_t\right] + b_c\right),$$
$$\boldsymbol{h}_t = \boldsymbol{o}_t \odot \tanh(\boldsymbol{c}_t), \quad (4)$$

$$\boldsymbol{\Phi} = \{\phi_{q,p}\}, p = 1,2,\cdots,n_{\text{in}}, q = 1,2,\cdots,n_{\text{out}}, \quad (6)$$

where the functions $\phi_{q,p}$ have trainable parameters.

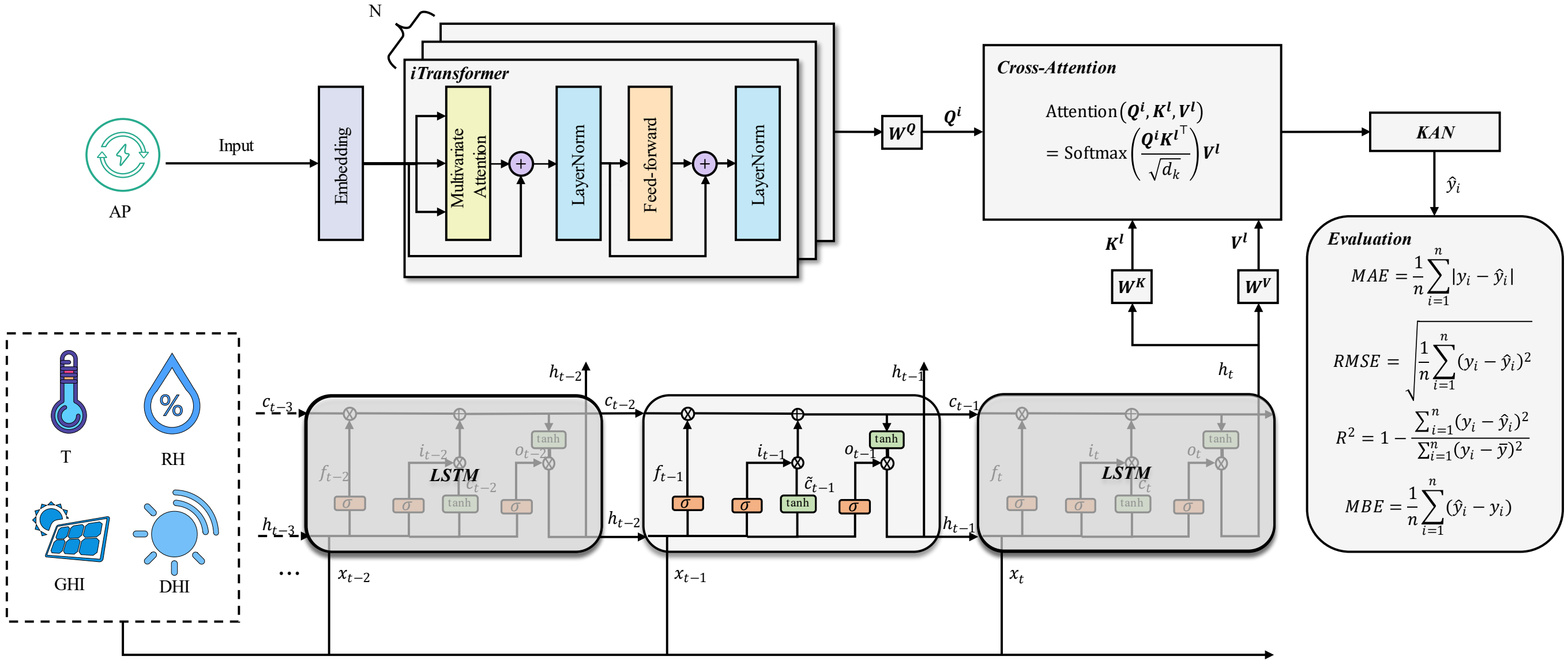

Fig. 1. The structure of proposed model.

where $\boldsymbol{x}_t, \boldsymbol{c}_t$, and $\boldsymbol{h}_t$ represent the current input, cell state, and output vectors, respectively, $\boldsymbol{c}_{t-1}$ and $\boldsymbol{h}_{t-1}$ are the previous cell state and output vector. The input, forget, and output gates produce outputs $\boldsymbol{i}_t$, $\boldsymbol{f}_t$, and $\boldsymbol{o}_t$, which regulate the flow of information. The weight matrices corresponding to the gates and memory cell are $\boldsymbol{W}_i, \boldsymbol{W}_f, \boldsymbol{W}_o$, and $\boldsymbol{W}_c$, with their associated biases $b_i, b_f, b_o$, and $b_c$. $\sigma(\cdot)$ is the sigmoid activation function, $\odot$ denotes the element-wise multiplication operation, and $\tanh(\cdot)$ is the hyperbolic tangent activation function.

*D. Kolmogorov–Arnold Networks*

KAN are a type of neural network architecture inspired by the Kolmogorov-Arnold representation theorem [19], offering a key distinction from traditional MLP. While MLPs apply fixed activation functions at the nodes (neurons), KAN introduces learnable activation functions on the edges (weights), allowing for more flexible and dynamic learning. The Kolmogorov-Arnold representation theorem states that any continuous multivariate function can be represented as a composition of univariate functions along with addition operations:

$$f(\boldsymbol{x}) = f(x_1, x_1, \cdots, x_n) = \sum_{q=1}^{2n+1} \Phi_q\left(\sum_{p=1}^{n} \phi_{q,p}(x_p)\right), \quad (5)$$

where $\phi_{q,p}(\cdot)$ represents univariate functions that transform each input variable $x_p$ defined as $\phi_{q,p}: [0,1] \to \mathbb{R}$ and $\Phi_q: \mathbb{R} \to \mathbb{R}$.

Since all the functions to be learned are univariate, [19] introduced B-spline curves to estimate each function and define a KAN layer as

Thus, Eq. (5) represents a two-layer KAN structure, where the first layer has $n_{\text{in}} = n$ and $n_{\text{out}} = 2n + 1$, and the second layer has $n_{\text{in}} = 2n + 1$ and $n_{\text{out}} = 1$. Similar to an MLP, the output of a KAN with $L$ layers and with the input $\boldsymbol{X}_0$ can be represented as:

$$\text{KAN}(\boldsymbol{X}_0) = (\boldsymbol{\Phi}_{L-1} \circ \boldsymbol{\Phi}_{L-2} \circ \cdots \circ \boldsymbol{\Phi}_1 \circ \boldsymbol{\Phi}_0)\boldsymbol{X}_0, \quad (7)$$

$$\boldsymbol{X}_{l+1} = \boldsymbol{\Phi}_l \boldsymbol{X}_l$$
$$= \underbrace{\begin{pmatrix} \phi_{l,1,1}(\cdot) & \phi_{l,1,2}(\cdot) & \cdots & \phi_{l,1,n_l}(\cdot) \\ \phi_{l,2,1}(\cdot) & \phi_{l,2,2}(\cdot) & \cdots & \phi_{l,2,n_l}(\cdot) \\ \vdots & \vdots & & \vdots \\ \phi_{l,n_{l+1},1}(\cdot) & \phi_{l,n_{l+1},2}(\cdot) & \cdots & \phi_{l,n_{l+1},n_l}(\cdot) \end{pmatrix}}_{\boldsymbol{\Phi}_l} \underbrace{\begin{pmatrix} x_{l,1} \\ x_{l,2} \\ \vdots \\ x_{l,n_l} \end{pmatrix}}_{\boldsymbol{X}_l}, \quad (8)$$

where $\boldsymbol{X}_l \in \mathbb{R}^{n_l}$ and $\boldsymbol{X}_{l+1} \in \mathbb{R}^{n_{l+1}}$ are the pre-activation vector and post-activation vector of function matrix corresponding to the $l^{th}$ KAN layer $\boldsymbol{\Phi}_l$, respectively. $x_{l,i}$ denotes the activation value of the $i^{th}$ neuron in the $l^{th}$ layer and $\phi_{l,j,i}(\cdot)$ is the activation function that connects $x_{l,i}$ and $x_{l+1,j}$.

*E. Proposed Model*

For multivariate-to-univariate forecasting tasks, the iTransformer may not fully perform as it operates primarily on the variable dimension. To address this, the input variables are divided into target variables and covariates. The target variable is encoded using the iTransformer, while the LSTM captures features from the covariates. Finally, cross-attention enhances the fusion of features between the target and covariates, and KAN maps these fused features to the desired forecast dimension. The structure of the proposed model is illustrated in Fig. 1.

## III. CASE STUDY

This section presents the experimental data and the performance of the proposed model in the PV power dataset across four seasons. All experiments were conducted on a Windows 10 system with an Intel Xeon Silver 4214 CPU @ 2.20GHz and an NVIDIA GeForce RTX 2080 Ti GPU with 12GB of video memory. The software environment consisted of Python 3.9 and PyTorch 2.4.1 and optuna 4.0.0 [21].

### A. Data selection and preprocessing

The PV power data used in this study were sourced from the Desert Knowledge Australia Solar Centre[1], specifically from Site 7 in Alice Springs, Australia (latitude: -23.76, longitude: 133.87). PV power data vary across seasons. To evaluate model performance under different conditions, the data were segmented into four seasons: Spring (September to November), Summer (December to February), Autumn (March to May), and Winter (June to August) , covering a period of three years with a time resolution of 1 hour. Missing data were filled using forward linear interpolation, while negative active power values were set to zero due to potential sensor errors. The data were split into training, validation, and test sets in an 8:1:1 ratio. TABLE I. provides the details of the PV system parameters.

TABLE I. PARAMETERS IN THE OBSERVED PV SYSTEM.

| PV System (Site 7) | Parameters |
|---|---|
| Array Rating | 6.96kW |
| Panel Rating | 73W |
| Number Of Panels | 96 |
| Array Area | 69.12 $m^2$ |
| Inverter Size/Type | 6 kW, Fronius Primo 6.0-1 |
| Array Tilt/Azimuth | Tilt=20, Azi=0 (Solar North) |

Based on the feature selection method from [22], the chosen input variables include active power ( AP, $kW$ ), historical temperature ( $T$, ℃ ), relative humidity (RH, % ), global horizontal irradiance (GHI, $Wh/m^2$), and diffuse horizontal irradiance (DHI, $Wh/m^2$), as shown in Fig. 2. Each input variable is normalized individually as

$$\hat{x} = \frac{x - \mu}{\sigma}, \tag{9}$$

where $x$ is the original data, $\mu$ is the mean of $x$ across the training data, $\sigma$ is the standard deviation of $x$ across the training data and $\hat{x}$ is the normalized data.

A sliding window approach is applied, using data from the past one day $\boldsymbol{X} = (AP, T, RH, GHI, DHI) \in \mathbb{R}^{5\times 24}$ to forecast the next one hour $y = (AP) \in \mathbb{R}^1$.

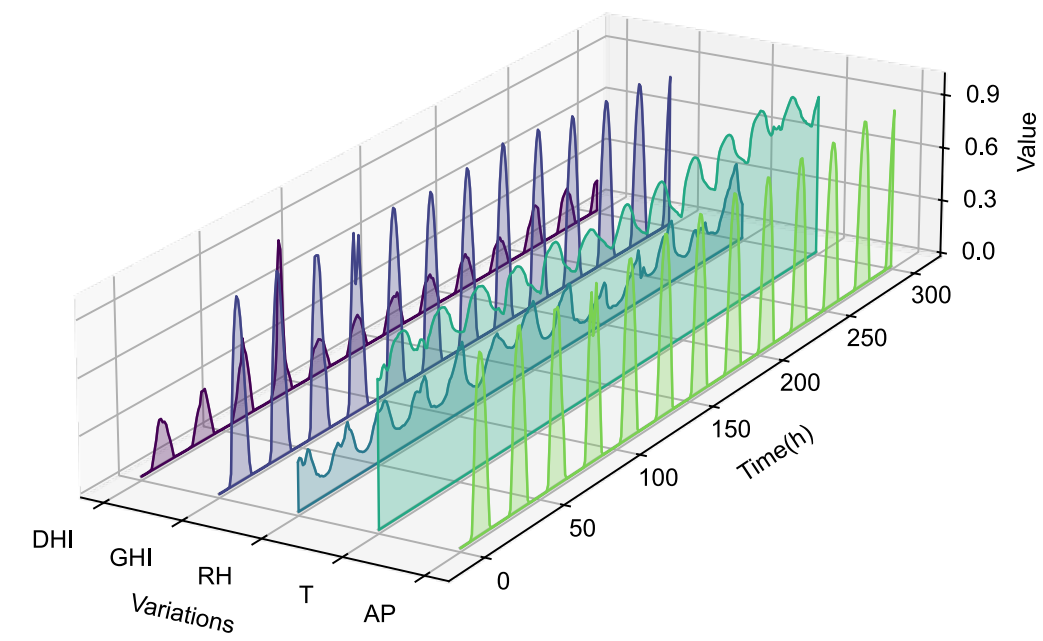

Fig. 2. Waterfall chart of input variables.

### B. Experimental setting

To demonstrate the effectiveness of the proposed model, comparisons were made between iTransformer, LSTM, and the proposed model. All models were trained according to the methods outlined in TABLE II. , with hyperparameter selections detailed in TABLE III. . To prevent overfitting, an early stopping strategy was employed, and Bayesian optimization with Optuna[2] was used to search for the optimal parameter combination within the parameter space.

TABLE II. UNIFORM CRITERIA OF TRAINING FOR MODELS.

| Model Training Setup | Values |
|---|---|
| Batch Size | 128 |
| Epoch | 150 |
| optimizer | Adam |
| Learning Rate | 0.0001 |
| Learning Rate Scheduling Strategy | ReduceLROnPlateau |

TABLE III. HYPERPARAMETER SETTINGS FOR MODELS.

| Hyperparameters | Optional Values |
|---|---|
| Embedding dimension of iTransformer $\boldsymbol{d_i}$ | {16, 32, 64, 128, 256} |
| Layer of iTransformer $\boldsymbol{l_i}$ | {1, 2, 3, 4, 5, 6} |
| Head of attention $\boldsymbol{h}$ | {2, 4, 6, 8, 12, 16} |
| Hidden dimension of LSTM $\boldsymbol{d_l}$ | {16, 32, 64, 128, 256} |
| Layer of LSTM $\boldsymbol{l_l}$ | {1, 2, 3, 4} |
| KAN layer | $\{n_{\text{in}} = d_i, n_{\text{out}} = 1\}$ |

### C. Evaluation Metrics

This study evaluates the forecasting performance of different models using mean absolute error (MAE), root mean square error (RMSE), the coefficient of determination ($R^2$), and mean bias error (MBE) [23]. Smaller values of MAE and RMSE are preferable, while an $R^2$ closer to 1 is ideal. Additionally, the MBE should be as close to 0 as possible for optimal performance. The metrics are given by

$$MAE = \frac{1}{n}\sum_{i=1}^{n}|y_i - \hat{y}_i|, \tag{10}$$

$$RMSE = \sqrt{\frac{1}{n}\sum_{i=1}^{n}(y_i - \hat{y}_i)^2}, \tag{11}$$

[1] https://dkasolarcentre.com.au/

[2] https://github.com/optuna/optuna

$$R^2 = 1 - \frac{\sum_{i=1}^{n}(y_i - \hat{y}_i)^2}{\sum_{i=1}^{n}(y_i - \bar{y})^2}, \quad (12)$$

$$MBE = \frac{1}{n}\sum_{i=1}^{n}(\hat{y}_i - y_i), \quad (13)$$

where $n$ represents the all number of forecasted samples, $\bar{y}$ denotes the average value of all observed target values, $\hat{y}_i$ is the $i^{th}$ forecasted target value, and $y_i$ is the corresponding observed target value.

### D. Results

The forecasting performance of the proposed model, alongside benchmark models iTransformer and LSTM, is evaluated across four seasons, as shown in TABLE IV. and Fig. 3. These metrics provide a comprehensive assessment of each model's accuracy, predictive precision, and bias when forecasting PV power under seasonally varying data characteristics.

In Spring, the proposed model achieves an MAE of 0.1335 and an RMSE of 0.3406, with an $R^2$ value of 0.9646, the highest among the models. These metrics indicate a strong correlation with observed values, underscoring the proposed model's accuracy in capturing seasonal PV data variability. The iTransformer model, in comparison, yields a slightly higher MAE of 0.1455 and RMSE of 0.3448, reflecting marginally lower accuracy, but it reaches the lowest MBE which indicates a slight tendency towards overestimation. The LSTM, with an MAE of 0.1448 and a higher RMSE of 0.3542, shows a noticeable MBE of 0.0387, suggesting a tendency toward positive bias.

During Summer, all models display enhanced forecast accuracy. The proposed model achieves its lowest seasonal RMSE of 0.1153, along with an $R^2$ of 0.9966. This performance highlights the proposed model's robustness in the relatively stable, high-sunlight conditions of Summer. While the iTransformer also performs well with an MAE of 0.0591, RMSE of 0.1316, and $R^2$ of 0.9956, the proposed model consistently outperforms it in terms of precision and accuracy. The LSTM model has a comparable MAE of 0.0533 but a slightly higher RMSE of 0.1278. Notably, all models display a small negative MBE, indicating slight underestimation, with the proposed model's bias being the smallest at -0.0047.

Autumn, characterized by increased data variability, poses a more complex forecasting scenario. The proposed model achieves high accuracy with an MAE of 0.0986, RMSE of 0.2574, and an $R^2$ of 0.9761, demonstrating its robustness against seasonal shifts. The iTransformer, with an MAE of 0.1073 and RMSE of 0.2805, shows slightly reduced performance relative to the proposed model. The LSTM model exhibits the lowest accuracy in Autumn, with an MAE of 0.1182, RMSE of 0.323, and the lowest $R^2$ value 0.9623, indicating its limited adaptability to Autumn's increased variability.

In Winter, the proposed model continues to achieve the best accuracy, with the lowest MAE of 0.0428 and RMSE of 0.1717, along with a high $R^2$ of 0.9913. These results reflect the model's stable forecasting performance even under Winter's lower sunlight conditions. Both iTransformer and LSTM models perform with slightly higher MAE values of 0.0468 and 0.0473, respectively, and increased RMSEs of 0.1806 and 0.1907, respectively, indicating reduced accuracy compared to the proposed model. The proposed model exhibits a modest positive MBE of 0.0126, suggesting a minor degree of overestimation, though this remains within acceptable limits.

Overall, the proposed model consistently outperforms iTransformer and LSTM across all seasonal datasets, as evidenced by lower MAE and RMSE values and higher $R^2$ scores. These metrics affirm the proposed model's capability to capture seasonal variations in PV power data more effectively than the benchmarks. While LSTM tends to exhibit greater bias across seasons, iTransformer performs relatively close to the proposed model and achieves the best MBE relatively but demonstrates reduced precision. In the summer, all models underestimated PV power output, likely due to their inability to accurately capture the peak solar irradiance conditions typical of this season. Conversely, during Spring, Autumn, and Winter, the models exhibited an overestimation of power generation, potentially stemming from biases in historical data and inherent limitations in how the models process seasonal variations.

The findings validate the proposed model's suitability for accurate and resilient PV power forecasting across diverse seasonal patterns.

TABLE IV. VALUES OF EVALUATION METRICS ACROSS FOUR SEASONS.

| Seasons | Models | MAE | RMSE | $R^2$ | MBE |
|---|---|---|---|---|---|
| Spring | Proposed | **0.1335** | **0.3406** | **0.9646** | 0.0006 |
| | iTransformer | 0.1455 | 0.3448 | 0.9637 | **0.0004** |
| | LSTM | 0.1448 | 0.3542 | 0.9617 | 0.0387 |
| Summer | Proposed | **0.0517** | **0.1153** | **0.9966** | **-0.0047** |
| | iTransformer | 0.0591 | 0.1316 | 0.9956 | -0.0062 |
| | LSTM | 0.0533 | 0.1278 | 0.9956 | -0.0052 |
| Autumn | Proposed | **0.0986** | **0.2574** | **0.9761** | 0.0258 |
| | iTransformer | 0.1073 | 0.2805 | 0.9716 | **0.0106** |
| | LSTM | 0.1182 | 0.323 | 0.9623 | 0.0373 |
| Winter | Proposed | **0.0428** | **0.1717** | **0.9913** | 0.0126 |
| | iTransformer | 0.0468 | 0.1806 | 0.9903 | **0.0032** |
| | LSTM | 0.0473 | 0.1907 | 0.9892 | 0.0115 |

## IV. CONCLUSION

Accurate PV power forecasting is vital for stable grid operation and integrating solar energy effectively, supporting sustainability and carbon neutrality goals. The proposed model significantly improves forecasting accuracy by combining iTransformer encoding for target variables with LSTM-based feature extraction specifically designed for covariates. This separation in feature extraction addresses the unique characteristics of the target and covariate variables, while cross-attention effectively captures the interactions across variable dimensions and temporal dimensions. Achieving superior MAE, RMSE, and $R^2$ metrics across all seasons, these results highlight the model's reliability for PV forecasting and its adaptability to diverse conditions.

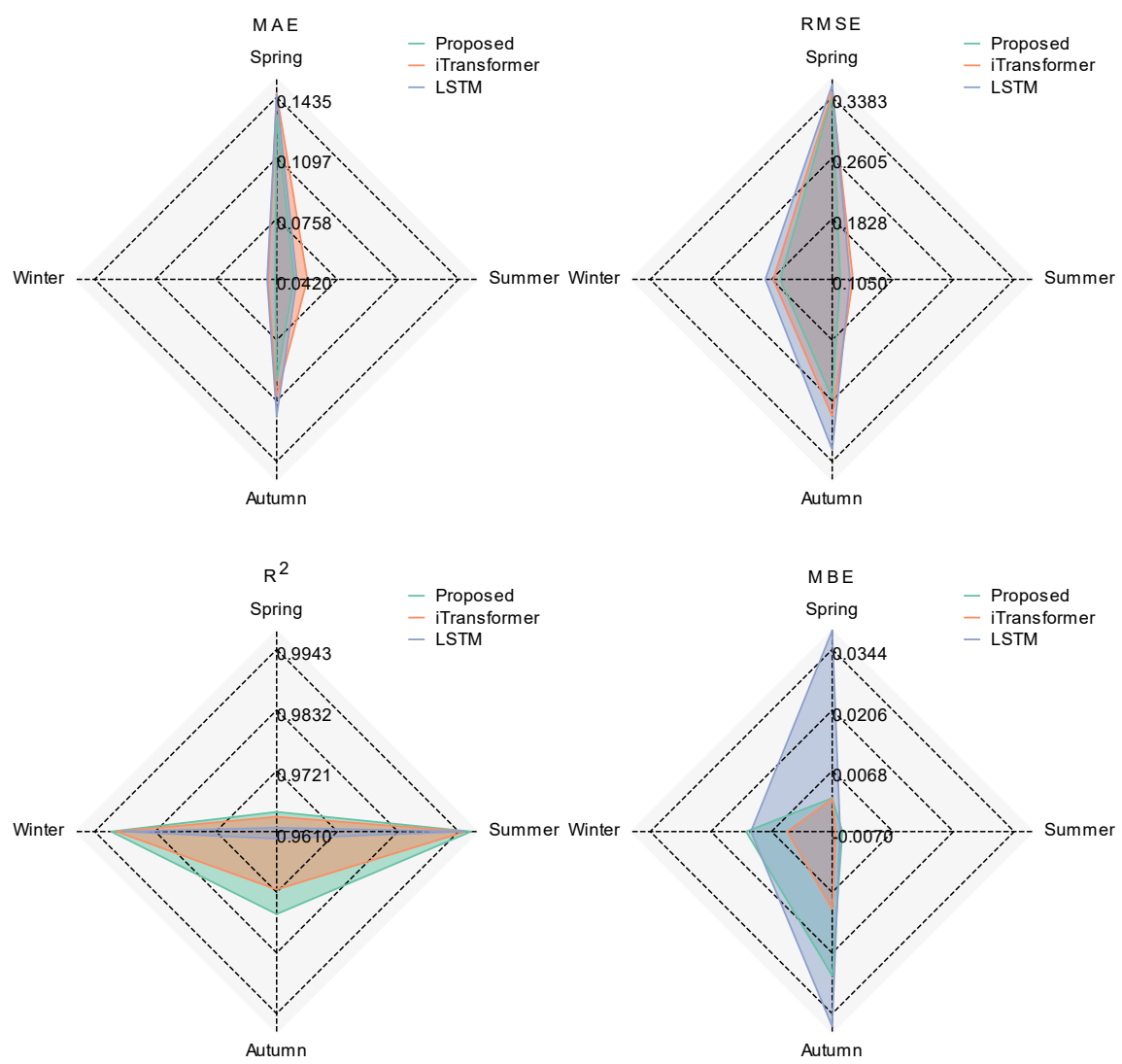

Fig. 3. Radar chart of evaluation metrics across four seasons.